\documentclass{article}

\usepackage{microtype}
\usepackage{graphicx}
\usepackage{booktabs} 
\usepackage{listings}
\usepackage{amsmath}
\usepackage{xcolor}
\usepackage{blindtext}
\usepackage{multirow}

\definecolor{codegreen}{rgb}{0,0.6,0}
\definecolor{codegray}{rgb}{0.5,0.5,0.5}
\definecolor{codepurple}{rgb}{0.58,0,0.82}
\definecolor{backcolour}{rgb}{0.95,0.95,0.95}

\lstdefinestyle{mystyle}{
    backgroundcolor=\color{backcolour},   
    commentstyle=\color{codegreen},
    keywordstyle=\color{magenta},
    numberstyle=\tiny\color{codegray},
    stringstyle=\color{codepurple},
    basicstyle=\ttfamily\footnotesize,
    breakatwhitespace=false,         
    breaklines=true,                 
    captionpos=b,                    
    keepspaces=true,                 
    numbersep=5pt,                  
    showspaces=false,                
    showstringspaces=false,
    showtabs=false,                  
    tabsize=2
}

\usepackage{url}

\usepackage{breakurl}
\usepackage[breaklinks]{hyperref}
 
\usepackage[accepted]{icml2020}

\begin{document}

\twocolumn[
\icmltitle{Megatron-LM: Training Multi-Billion Parameter Language Models Using Model Parallelism}

\begin{icmlauthorlist}
\icmlauthor{Mohammad Shoeybi}{equal,to}
\icmlauthor{Mostofa Patwary}{equal,to}
\icmlauthor{Raul Puri}{equal,to}
\icmlauthor{Patrick LeGresley}{to}
\icmlauthor{Jared Casper}{to}
\icmlauthor{Bryan Catanzaro}{to}
\end{icmlauthorlist}
\icmlaffiliation{to}{NVIDIA}
\icmlaffiliation{equal}{Equal contribution}
\icmlcorrespondingauthor{Mohammad Shoeybi}{mshoeybi@nvidia.com}

\icmlkeywords{Machine Learning}

\vskip 0.3in

]

\printAffiliationsAndNotice{}

\begin{abstract}

Recent work in language modeling demonstrates that training large transformer models advances the state of the art in Natural Language Processing applications. However, very large models can be quite difficult to train due to memory constraints. In this work, we present our techniques for training very large transformer models and implement a simple, efficient intra-layer model parallel approach that enables training transformer models with billions of parameters. Our approach does not require a new compiler or library changes, is orthogonal and complimentary to pipeline model parallelism, and can be fully implemented with the insertion of a few communication operations in native PyTorch. We illustrate this approach by converging transformer based models up to 8.3 billion parameters using 512 GPUs. We sustain 15.1 PetaFLOPs  across the entire application with 76\% scaling efficiency when compared to a strong single GPU baseline that sustains 39 TeraFLOPs, which is 30\% of peak FLOPs. To demonstrate that large language models can further advance the state of the art (SOTA), we train an 8.3 billion parameter transformer language model similar to GPT-2 and a 3.9 billion parameter model similar to BERT. We show that careful attention to the placement of layer normalization in BERT-like models is critical to achieving increased performance as the model size grows. Using the GPT-2 model we achieve SOTA results on the WikiText103 (10.8 compared to SOTA perplexity of 15.8) and LAMBADA (66.5\% compared to SOTA accuracy of 63.2\%) datasets. Our BERT model achieves SOTA results on the RACE dataset (90.9\% compared to SOTA accuracy of 89.4\%).
  
\end{abstract}
\section{Introduction}

Natural Language Processing (NLP) is advancing quickly in part due to an increase in available compute and dataset size. The abundance of compute and data enables training increasingly larger language models via unsupervised pretraining \citep{devlin2018bert,Radford2019GPT2}. Empirical evidence indicates that larger language models are dramatically more useful for NLP tasks such as article completion, question answering, and natural language inference \cite{ALBERT2019,T5}. By finetuning these pretrained language models on downstream natural language tasks, one can achieve state of the art results as shown in recent work \cite{devlin2018bert, ELMo, Howard2018ULMFIT, Radford2018GPT, Radford2017Sentiment, Le2016seq2seqtransfer, roberta,transformerxl,xlnet,mtdnn,ALBERT2019}. 
    
As these models become larger, they exceed the memory limit of modern processors, and require additional memory management techniques such as activation checkpointing \citep{activation_checkpointing}. Widely used optimization algorithms such as ADAM require additional memory per parameter to store momentum and other optimizer state, which reduces the size of models that can be effectively trained. Several approaches to model parallelism overcome this limit by partitioning the model such that the weights and their associated optimizer state do not need to reside concurrently on the processor. For example, GPipe \cite{GPipe} and Mesh-Tensorflow \cite{mesh_tf} provide frameworks for model parallelism of different kinds. However, they require rewriting the model, and rely on custom compilers and frameworks that are still under development.
    
In this work,  we implement a simple and efficient model parallel approach using intra-layer model-parallelism. We exploit the inherent structure in transformer based language models to make a simple model-parallel implementation that trains efficiently in PyTorch, with no custom C++ code or compiler required. This approach is orthogonal to pipeline-based model parallelism as advocated by approaches such as GPipe~\cite{GPipe}. 
    
\begin{figure}
\begin{center}
 \includegraphics[scale=0.18]{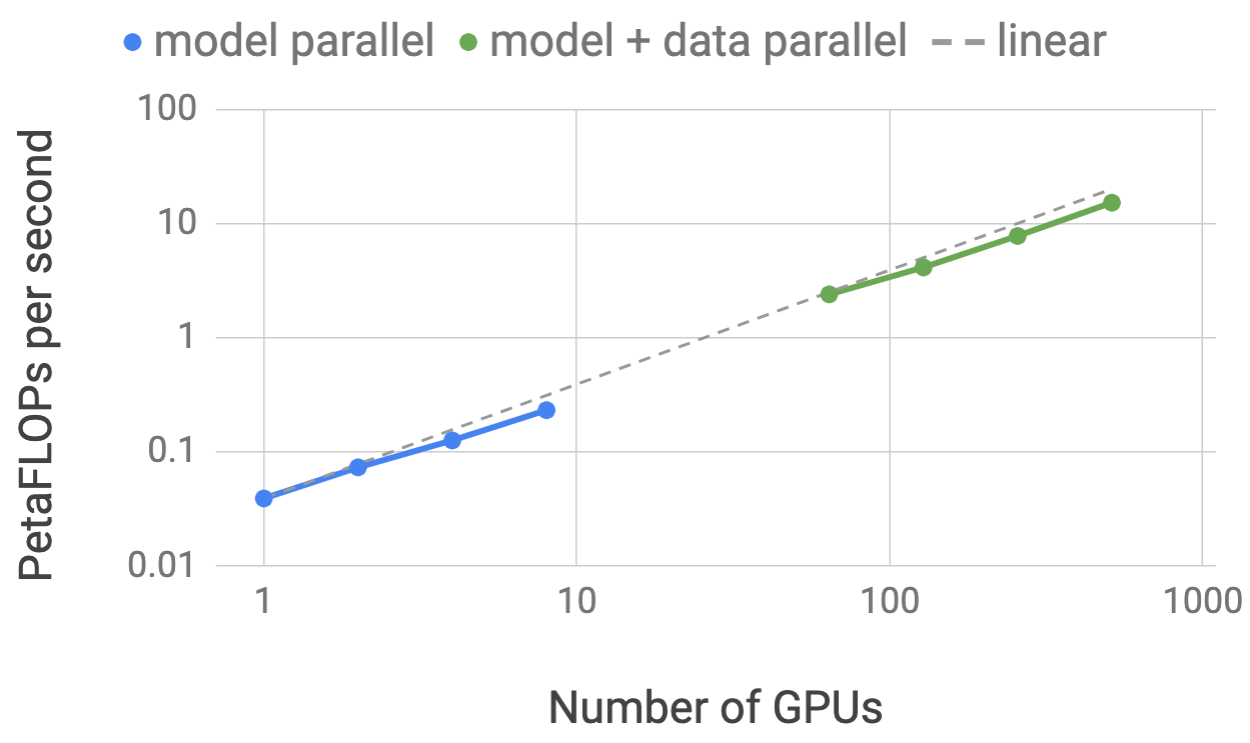}
 \caption{Model (blue) and model+data (green) parallel FLOPS as a function of number of GPUs. Model parallel (blue): up to 8-way model parallel weak scaling with approximately 1 billion parameters per GPU (e.g. 2 billion for 2 GPUs and 4 billion for 4 GPUs). Model+data parallel (green): similar configuration as model parallel combined with 64-way data parallel.}
 \label{fig:scale_line}
\end{center}
\end{figure}

To demonstrate the scalability of our approach, we establish a baseline by training a model of 1.2 billion parameters on a single NVIDIA V100 32GB GPU, that sustains 39 TeraFLOPs. This is 30\% of the theoretical peak FLOPS for a single GPU as configured in a DGX-2H server, and is thus a strong baseline. Scaling the model to 8.3 billion parameters on 512 GPUs with 8-way model parallelism, we achieve up to 15.1 PetaFLOPs per second sustained over the entire application. This is 76\% scaling efficiency compared to the single GPU case. Figure \ref{fig:scale_line} shows more detailed scaling results.
    
To analyze the effect of model size scaling on accuracy, we train both left-to-right GPT-2 \cite{Radford2019GPT2} language models as well as BERT \cite{devlin2018bert} bidirectional transformers and evaluate them on several downstream tasks. We show that the existing BERT architecture results in model degradation as the size increases. We overcome this challenge by rearranging the layer normalization and residual connection in the transformer layers and show that with this change, results for the downstream tasks  on development sets improve monotonically as the model size increases. In addition, we show that our models achieve test set state of the art (SOTA) results on WikiText103, cloze-style prediction accuracy on LAMBADA, and reading comprehension RACE datasets.

In summary, our contributions are as follows:
\begin{itemize}
    \item We implement a simple and efficient model parallel approach by making only a few targeted modifications to an existing PyTorch transformer implementation.
    \item We perform an in-depth empirical analysis of our model and data parallel technique and demonstrate up to 76\% scaling efficiency using 512 GPUs.
    \item We show that careful attention to the placement of layer normalization in BERT-like models is critical to achieving increased accuracies as the model grows.
    \item We demonstrate that scaling the model size results in improved accuracies for both GPT-2 (studied up to 8.3 billion parameters) and BERT (studied up to 3.9B parameters) models.
    \item We showcase that our models achieve state of the art results on test sets: perplexity on WikiText103 (10.8 ppl), accuracy on LAMBADA (66.5\%), and accuracy on RACE (90.9\%).
    \item We open source our code along with the training and evaluation pipelines at \url{https://github.com/NVIDIA/Megatron-LM}
\end{itemize}

\section{Background and Challenges}
\subsection{Neural Language Model Pretraining} 

Pretrained language models have become an indispensable part of NLP researchers' toolkits. Leveraging large corpus pretraining to learn robust neural representations of language is an active area of research that has spanned the past decade. Early examples of pretraining and transferring neural representations of language demonstrated that pretrained word embedding tables improve downstream task results compared to word embedding tables learned from scratch~\cite{Mikolov2013,Pennington2014,Turian2010}. 
Later work advanced research in this area by learning and transferring neural models that capture contextual representations of words~\citep{context2vec, Cove, ELMo, Radford2017Sentiment, Radford2019GPT2}. Recent parallel work \citep{Le2016seq2seqtransfer, Howard2018ULMFIT, Radford2018GPT, devlin2018bert, roberta, transformerxl,xlnet,mtdnn,ALBERT2019} further builds upon these ideas by not just transferring the language model to extract contextual word representations, but by also finetuning the language model in an end to end fashion on downstream tasks. Through these works, the state of the art has advanced from transferring just word embedding tables to transferring entire multi-billion parameter language models. This progression of methods has necessitated the need for hardware, systems techniques, and frameworks that are able to operate efficiently at scale and satisfy increasing computational needs. Our work aims to provide the tools necessary to take another step forward in this trend.

\subsection{Transformer Language Models and Multi-Head Attention} 

Current work in NLP trends towards using {\it transformer} models~\cite{Transformer} due to their superior accuracy and compute efficiency. The original transformer formulation was designed as a machine translation architecture that transforms an input sequence into another output sequence using two parts, an {\it Encoder} and {\it Decoder}. However, recent work leveraging transformers for language modeling such as BERT \cite{devlin2018bert} and GPT-2 \cite{Radford2019GPT2} use only the {\it Encoder} or {\it Decoder} depending on their needs. This work explores both a decoder architecture, GPT-2, and an encoder architecture, BERT. 

\begin{figure}[t!]
\vspace{-2mm}
\begin{center}
  \includegraphics[scale=0.24]{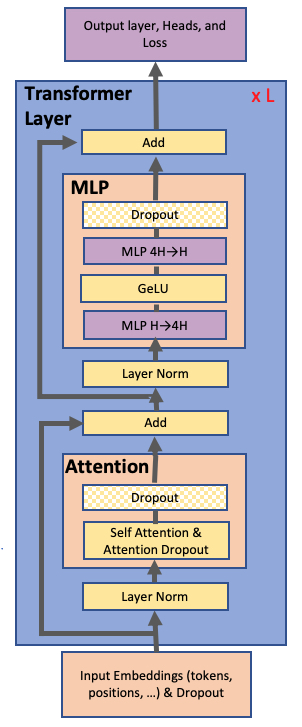}
  \vspace{-2mm}
  \caption{Transformer Architecture. Purple blocks correspond to fully connected layers. Each blue block represents a single transformer layer that is replicated N times.}
  \label{fig:gpt2-transformer}
\end{center}
\vspace{-6mm}
\end{figure}

Figure \ref{fig:gpt2-transformer} shows a schematic diagram of the model we used. We refer the reader to prior work for a detailed description of the model architecture~\citep{Transformer,devlin2018bert,Radford2019GPT2}. It is worthwhile to mention that both GPT-2 and BERT use GeLU \citep{gelu} nonlinearities and layer normalization \citep{layernorm} to the input of the multi-head attention and feed forward layers, whereas the original transformer \citep{Transformer} uses ReLU nonlinearities and applies layer normalization to outputs.

\subsection{Data and Model Parallelism in Deep Learning} 

There are two central paradigms for scaling out deep neural network training to numerous hardware accelerators: data parallelism \cite{DataParallel1990} where a training minibatch is split across multiple workers, and model parallelism in which the memory usage and computation of a model is distributed across multiple workers. By increasing the minibatch size proportionally to the number of available workers (i.e. \emph{weak scaling}), one observes near linear scaling in training data throughput. However, large batch training introduces complications into the optimization process that can result in reduced accuracy or longer time to convergence, offsetting the benefit of increased training throughput \cite{LargeBatch2016}. Further research \cite{Goyal2017, Lars2017, Lamb2019} has developed techniques to mitigate these effects and drive down the training time of large neural networks. To scale out training even further, parallel work \cite{activation_checkpointing} has combined data parallelism with activation checkpointing: recomputing activations in the backward pass without storing them in the forward pass to reduce memory requirements. 

However, these techniques have one fundamental limitation in the problem size they can tackle: the model must fit entirely on one worker. With language models of increasing size and complexity like BERT and GPT-2, neural networks have approached the memory capacity of modern hardware accelerators. One solution to this problem is to employ parameter sharing to reduce the memory footprint of the model \cite{ALBERT2019}, but this limits the overall capacity of the model. Our approach is to utilize model parallelism to split the model across multiple accelerators. This not only alleviates the memory pressure, but also increases the amount of parallelism independently of the microbatch size.

Within model parallelism, there are two further paradigms: layer-wise pipeline parallelism, and more general distributed tensor computation. In pipeline model parallelism, groups of operations are performed on one device before the outputs are passed to the next device in the pipeline where a different group of operations are performed. Some approaches \cite{PipeDream2018, DualPipe2018} use a parameter server \cite{ParamServer2014} in conjunction with pipeline parallelism. However these suffer from inconsistency issues. The GPipe framework for TensorFlow \cite{GPipe} overcomes this inconsistency issue by using synchronous gradient decent. This approach requires additional logic to handle the efficient pipelining of these communication and computation operations, and suffers from pipeline bubbles that reduce efficiency, or changes to the optimizer itself which impact accuracy.

Distributed tensor computation is an orthogonal and more general approach that partitions a tensor operation across multiple devices to accelerate computation or increase model size. FlexFlow \cite{FlexFlow2018}, a deep learning framework orchestrating such parallel computation, provides a method to pick the best parallelization strategy. Recently, Mesh-TensorFlow \cite{mesh_tf} introduced a language for specifying a general class of distributed tensor computations in TensorFlow \cite{tensorflow2015-whitepaper}. The parallel dimensions are specified in the language by the end user and the resulting graph is compiled with proper collective primitives. We utilize similar insights to those leveraged in Mesh-TensorFlow and exploit parallelism in computing the transformer's attention heads to parallelize our transformer model. However, rather than implementing a framework and compiler for model parallelism, we make only a few targeted modifications to existing PyTorch transformer implementations. Our approach is simple, does not require any new compiler or code re-writing, and can be fully implemented by inserting a few simple primitives, as described in the next section. 
\section{Model Parallel Transformers} 
\label{model_par}

We take advantage of the structure of transformer networks to create a simple model parallel implementation by adding a few synchronization primitives. A transformer layer consists of a self attention block followed by a two-layer, multi-layer perceptron (MLP) as shown in Figure \ref{fig:gpt2-transformer}. We introduce model parallelism in both of these blocks separately.

 We start by detailing the MLP block. The first part of the block is a GEMM followed by a GeLU nonlinearity:
 \begin{equation}
    Y = \textrm{GeLU}(XA)
    \label{eq:mlp-firstgemm}
 \end{equation}
 One option to parallelize the GEMM is to split the weight matrix $A$ along its rows and input $X$ along its columns as:
 \begin{equation}
    X = [X_1, X_2], \
    A=\begin{bmatrix}
        A_1 \\
        A_2
    \end{bmatrix}.
 \end{equation}
 This partitioning will result in $Y = \textrm{GeLU}(X_1A_1 + X_2A_2)$. Since GeLU is a nonlinear function, $\textrm{GeLU}(X_1A_1 + X_2A_2)\neq \textrm{GeLU}(X_1A_1)+\textrm{GeLU}(X_2A_2)$ and this approach will require a synchronization point before the GeLU function. 
 
 Another option is to split $A$ along its columns $A=[A_1, A_2]$. This partitioning allows the GeLU nonlinearity to be independently applied to the output of each partitioned GEMM:
 \begin{equation}
    [Y_1, Y_2]= [\textrm{GeLU}(XA_1), \textrm{GeLU}(XA_2)] 
 \end{equation}
 This is advantageous as it removes a synchronization point. Hence, we partition the first GEMM in this column parallel fashion and split the second GEMM along its rows so it takes the output of the GeLU layer directly without requiring any communication as shown in Figure \ref{fig:model-para-mlp}a. The output of the second GEMM is then reduced across the GPUs before passing the output to the dropout layer. This approach splits both GEMMs in the MLP block across GPUs and requires only a single all-reduce operation in the forward pass ($g$ operator) and a single all-reduce in the backward pass ($f$ operator). These two operators are conjugates of each other and can be implemented in PyTorch with only a few lines of code. As an example, the implementation of the $f$ operator is provided below:

\begin{figure}[t!]
%\vspace{-2mm}
\begin{center}
  \includegraphics[scale=0.15]{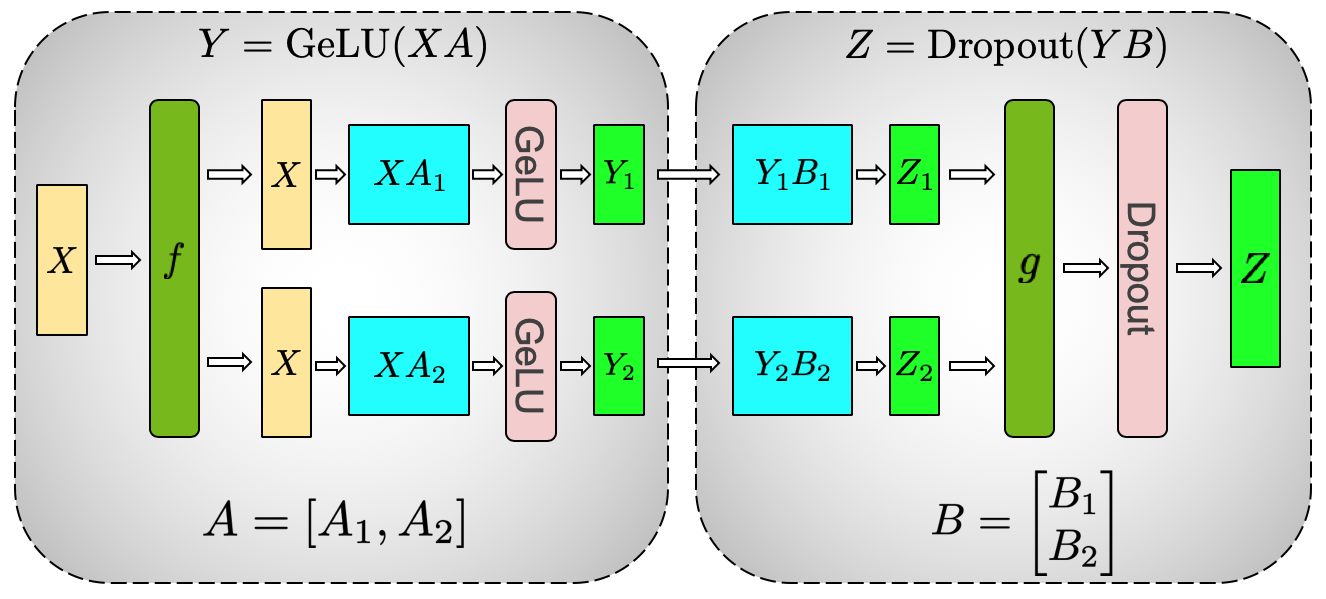}
\\
(a) MLP
\\
  \includegraphics[scale=0.2]{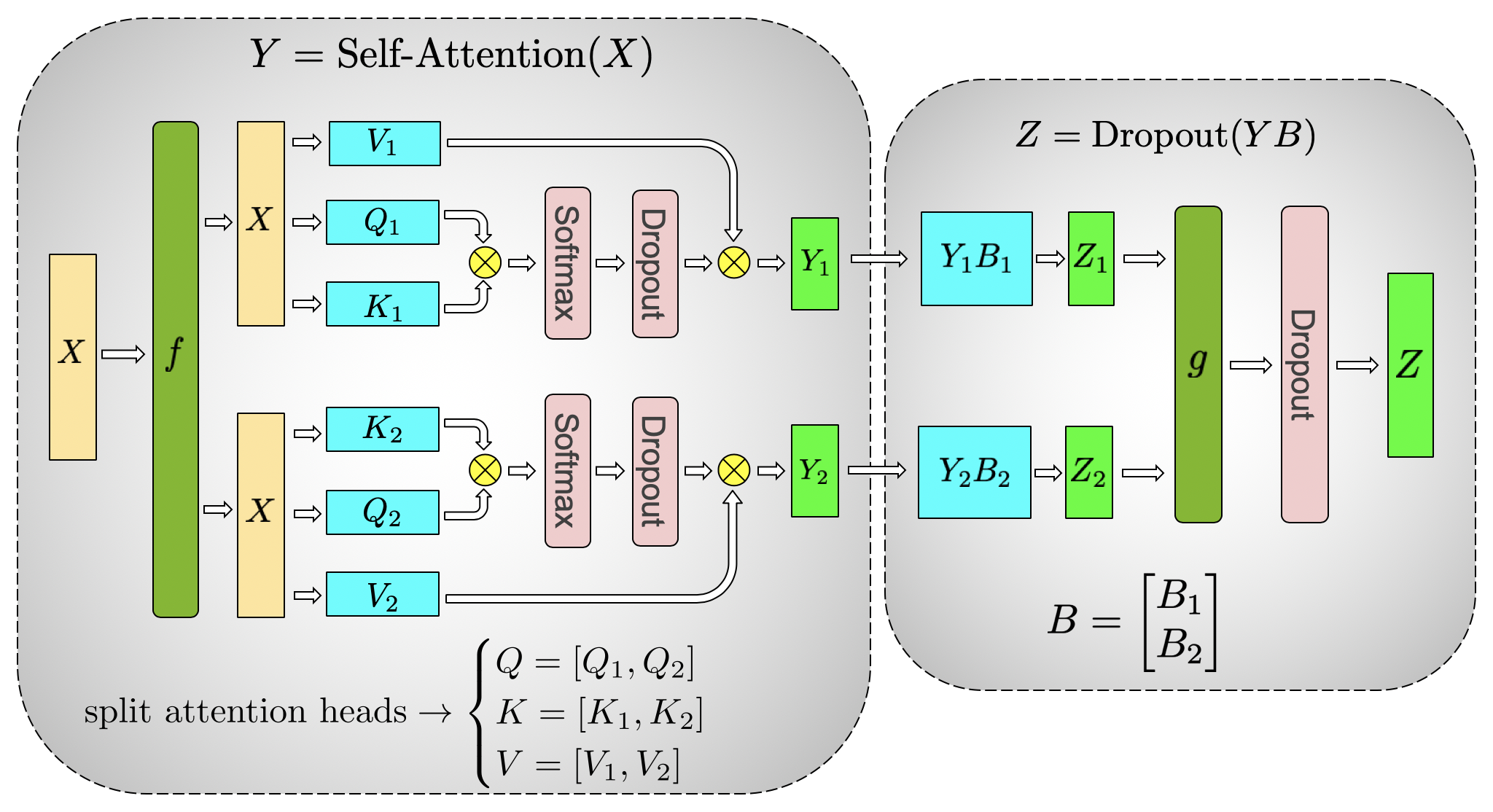}
\\
(b) Self-Attention
\\
  \vspace{-2mm}
  \caption{Blocks of Transformer with Model Parallelism. $f$ and $g$ are conjugate. $f$ is an identity operator in the forward pass and all reduce in the backward pass while $g$ is an all reduce in the forward pass and identity in the backward pass.}
  \label{fig:model-para-mlp}
\end{center}
\vspace{-6mm}
\end{figure}

\begin{lstlisting}[language=Python, caption={Implementation of $f$ operator. $g$ is similar to $f$ with identity in the backward and all-reduce in the forward functions.}]
class f(torch.autograd.Function):
    def forward(ctx, x):
        return x
    def backward(ctx, gradient):
        all_reduce(gradient)
        return gradient
\end{lstlisting}

\begin{figure}
\begin{center}
 \includegraphics[scale=0.22]{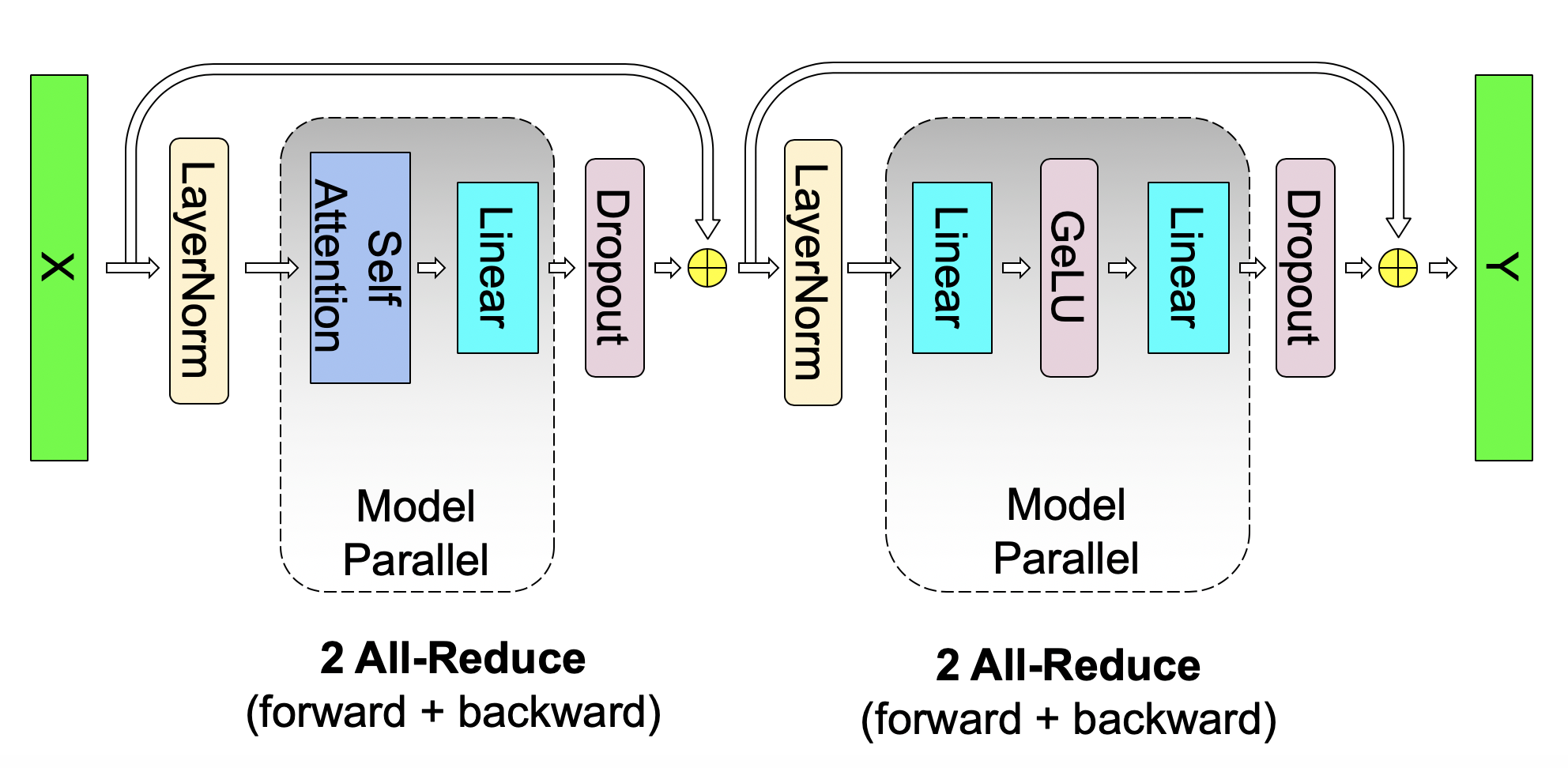}
 \caption{Communication operations in a transformer layer. There are 4 total communication operations in the forward and backward pass of a single model parallel transformer layer.}
 \label{fig:gpt2-transformerpasses}
 \vspace{-6mm}
\end{center}
\end{figure}

As shown in Figure \ref{fig:model-para-mlp}b, for the self attention block we exploit inherent parallelism in the multihead attention operation, partitioning the GEMMs associated with key ($K$), query ($Q$), and value ($V$) in a column parallel fashion such that the matrix multiply corresponding to each attention head is done locally on one GPU. This allows us to split per attention head parameters and workload across the GPUs, and doesn’t require any immediate communication to complete the self-attention. The subsequent GEMM from the output linear layer (after self attention) is parallelized along its rows and takes the output of the parallel attention layer directly, without requiring communication between the GPUs. This approach for both the MLP and self attention layer fuses groups of two GEMMs, eliminates a synchronization point in between, and results in better scaling. This enables us to perform all GEMMs in a simple transformer layer using only two all-reduces in the forward path and two in the backward path (see Figure \ref{fig:gpt2-transformerpasses}).

The transformer language model has an output embedding with the dimension of hidden-size ($H$) times vocabulary-size ($v$). Since the vocabulary size is on the order of tens of thousands of tokens for modern language models (for example, GPT-2 used a vocabulary size of 50,257), it is beneficial to parallelize the output embedding GEMM. However, in transformer language models, the output embedding layer shares weights with the input embedding, requiring modifications to both. We parallelize the input embedding weight matrix $E_{H\times v}$ along the vocabulary dimension $E=[E_1, E_2]$ (column-wise). Since each partition now only contains a portion of the embedding table, an all-reduce ($g$ operator) is required after the input embedding. For the output embedding, one approach is to perform the parallel GEMM $[Y_1, Y_2] = [XE_1, XE_2]$ to obtain the logits, add an all-gather $Y=\textrm{all-gather}([Y_1, Y_2])$, and send the results to the cross-entropy loss function. However, for this case, the all-gather will communicate $b\times s\times v$ elements ($b$ is the batch-size and $s$ is the sequence length) which is huge due to vocabulary size being large. To reduce the communication size, we fuse the output of the parallel GEMM $[Y_1, Y_2]$ with the cross entropy loss which reduces the dimension to $b \times s$. Communicating scalar losses instead of logits is a huge reduction in communication that improves the efficiency of our model parallel approach.

Much of our model parallel approach can be characterized as techniques aimed at reducing communication and keeping the GPUs compute bound.
Rather than having one GPU compute part of the dropout, layer normalization, or residual connections and broadcast the results to other GPUs, we choose to duplicate the computation across GPUs. Specifically, we maintain duplicate copies of layer normalization parameters on each GPU, and take the output of the model parallel region and run dropout and residual connection on these tensors before feeding them as input to the next model parallel regions. To optimize the model we allow each model parallel worker to optimize its own set of parameters. Since all values are either local to or duplicated on a GPU, there is no need for communicating updated parameter values in this formulation.

We present further details about the hybrid model and data parallelism and handling random number generation in Appendix \ref{sec:modelpar:supp} for reference. In summary, our approach as described above is simple to implement, requiring only a few extra all-reduce operations added to the forward and backward pass. It does not require a compiler, and is orthogonal and complementary to the pipeline model parallelism advocated by approaches such as \cite{GPipe}.

\section{Setup} 
\label{language_modeling}
Pretrained language understanding models are central tasks in natural language processing and language understanding. There are several formulations of language modeling. In this work we focus on GPT-2 \cite{Radford2019GPT2}, a left-to-right generative transformer based language model, and BERT \cite{devlin2018bert}, a bi-directional transformer model based on language model masking. We explain our configurations for these models in the following section and refer to the original papers for more details.

\subsection{Training Dataset}

To collect a large diverse training set with longterm dependencies we aggregate several of the largest language modeling datasets. We create an aggregate dataset consisting of Wikipedia \citep{devlin2018bert}, CC-Stories \citep{ccstories}, RealNews \citep{grover}, and OpenWebtext \cite{Radford2019GPT2}. To avoid training set leakage into our downstream tasks we remove the Wikipedia articles present in the WikiText103 test set \citep{wikitext}. We  also remove unnecessary newlines from the CC-Stories corpus introduced by preprocessing artifacts. For BERT models we include BooksCorpus \cite{BooksCorpus} in the training dataset, however, this dataset is excluded for GPT-2 trainings as it overlaps with LAMBADA task.

We combined all the datasets and then filtered out all the documents with content length less than 128 tokens from the aggregated dataset. Since similar content might appear multiple times in the aggregated datasets, we used locality-sensitive hashing (LSH) to deduplicate content with a jaccard similarity greater than 0.7. The resulting aggregate corpus contains 174 GB of deduplicated text.

\subsection{Training Optimization and Hyperparameters}
To train our models efficiently we utilize mixed precision training with dynamic loss scaling to take advantage of the V100's Tensor Cores \citep{MPTraining, AutoLossScale}. We start by initializing our weights $W$ with a simple normal distribution $W \sim \mathcal{N}(0, 0.02)$. We then scale weights immediately before residual layers by $\frac{1}{\sqrt{2N}}$ where N is the number of transformer layers comprised of self attention and MLP blocks. For our optimizer we utilize  Adam  \citep{Adam} with weight decay \citep{loshchilov2018decoupled} $\lambda = 0.01$. 
Additionally, we use global gradient norm clipping of 1.0 to improve the stability of training large models. In all cases, a dropout of 0.1 is used.
 Lastly, to better manage our memory footprint we utilize activation checkpointing \citep{activation_checkpointing} after every transformer layer.
 
%\subsubsection{GPT-2}

For GPT-2 models, all training is performed with sequences of 1024 subword units at a batch size of 512 for 300k iterations. Our learning rate of 1.5e-4 utilizes a warmup period of 3k iterations before following a single cycle cosine decay over the remaining 297k iterations. We stop the decay at a minimum learning rate of 1e-5.

%\subsubsection{BERT}
For BERT models, we largely follow the training process described in \cite{ALBERT2019}. We use the original BERT dictionary with vocab size of 30,522. In addition, we replace the next sentence prediction head with sentence order prediction as suggested by \cite{ALBERT2019} and use whole word n-gram masking of \cite{SpanBERT2019}. For all cases, we set the batch size to 1024 and use a learning rate of 1.0e-4 warmed up over 10,000 iterations and decayed linearly over 2 million iterations. Other training parameters are kept the same as \cite{devlin2018bert}.

\section{Experiments}

All of our experiments use up to 32 DGX-2H servers (a total of 512 Tesla V100 SXM3 32GB GPUs). Our infrastructure is optimized for multi-node deep learning applications, with 300 GB/sec bandwidth between GPUs inside a server via NVSwitch and 100 GB/sec of interconnect bandwidth between servers using 8 InfiniBand adapters per server.

\subsection{Scaling Analysis}
To test the scalability of our implementation, we consider GPT-2 models with four sets of parameters detailed in Table \ref{tab:params_scaling_studies}. To have consistent GEMM sizes in the self attention layer, the hidden size per attention head is kept constant at 96 while the number of heads and layers are varied to obtain configurations ranging from 1 billion to 8 billion parameters. The configuration with 1.2 billion parameters fits on a single GPU whereas the 8 billion parameter model requires 8-way model parallelism (8 GPUs). The original vocabulary size was 50,257, however, to have efficient GEMMs for the logit layer, it is beneficial for the per-GPU vocabulary size to be a multiple of 128. Since we study up to 8-way model parallelism, we pad the vocabulary such that it is divisible by $128\times 8=1024$, resulting in a padded vocabulary size of 51,200.  We study both model and model+data parallel scaling. For the model parallel scaling, a fixed batch size of 8 is used across all configurations. Data parallel scaling is necessary for training many state of the art models which typically use a much larger global batch size. To this end, for the model+data parallel cases we fix the global batch size to 512 for all experiments which corresponds to 64-way data parallelism. 

\begin{table}
\footnotesize
\begin{center}
\vspace{-3mm}
\caption{Parameters used for scaling studies. Hidden size per attention head is kept constant at 96.}
\vskip 0.15in
\label{tab:params_scaling_studies}
\begin{tabular}{c@{\hskip3pt}|@{\hskip3pt}c@{\hskip3pt}|@{\hskip3pt}c@{\hskip3pt}|@{\hskip3pt}c@{\hskip3pt}|@{\hskip3pt}c@{\hskip3pt}|@{\hskip3pt}c}
\hline \hline
 &  & Number & Number & Model & Model \\
Hidden & Attention & of & of & parallel & +data \\
Size & heads & layers & parameters & GPUs & parallel \\
&  & & (billions) & & GPUs \\ \hline
1536 &	16	& 40	& 1.2	& 1	& 64 \\
1920	& 20	&54	&2.5&	2&	128 \\
2304	&24	&64	&4.2	&4	&256 \\
3072	&32	& 72	&8.3	&8	&512 \\ \hline \hline
\end{tabular}
\end{center}
\vskip -0.1in
\vspace{-3mm}
\end{table}

\subsubsection{Model and Data Parallelism}
Throughout this section, we will showcase weak scaling with respect to the model parameters for both model parallel and model+data parallel cases. Weak scaling is typically done by scaling the batch-size, however, this approach does not address training large models that do not fit on a single GPU and it leads to training convergence degradation for large batch sizes. In contrast, here we use weak scaling to train larger models that were not possible otherwise. The baseline for all the scaling numbers is the first configuration (1.2 billion parameters) in Table \ref{tab:params_scaling_studies} running on a single GPU. This is a strong baseline as it achieves 39 TeraFLOPS during the overall training process, which is 30\% of the theoretical peak FLOPS for a single GPU in a DGX-2H server. 

\begin{figure}
%\vspace{-2mm}
\begin{center}
  \includegraphics[width=\columnwidth]{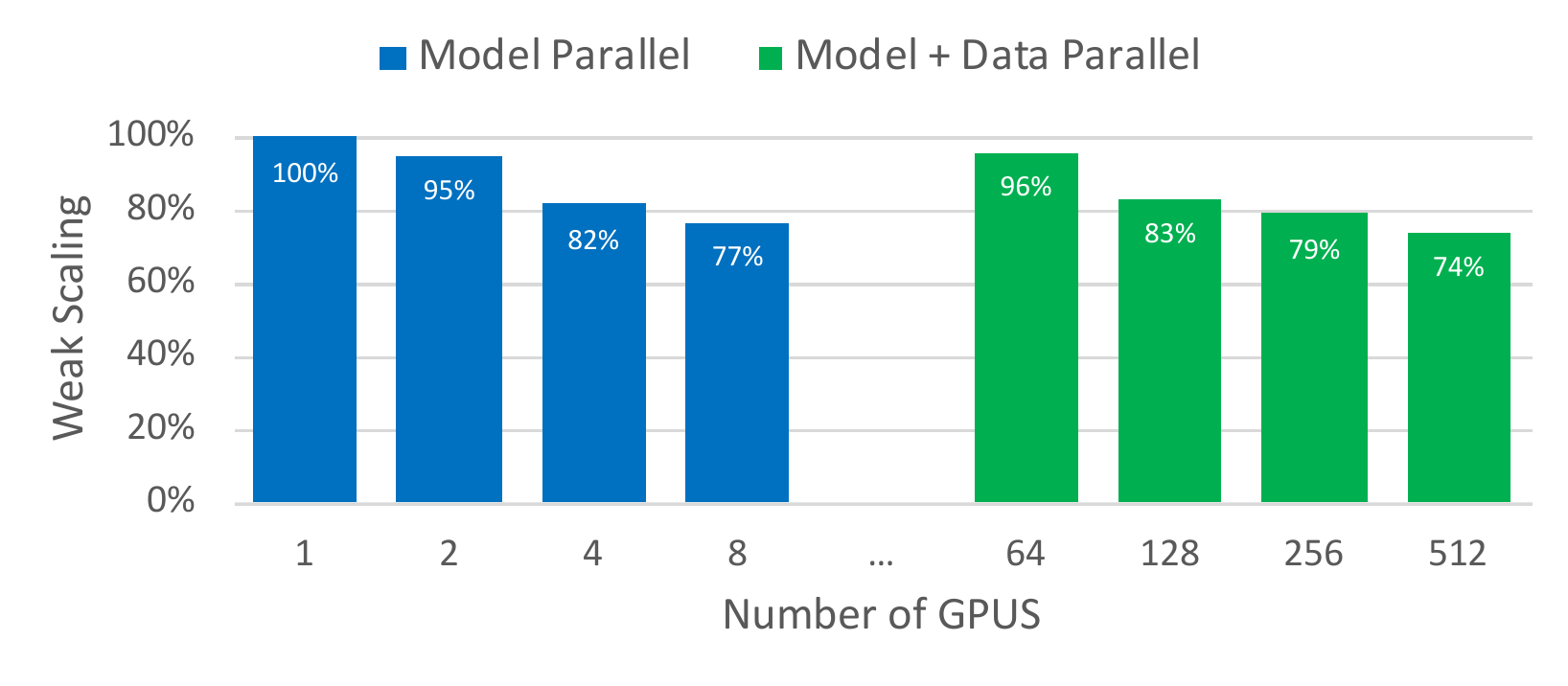}
  \vspace{-1cm}
  \caption{Model and model + data parallel weak scaling efficiency as a function of the number of GPUs.}
  \label{fig:blue_green_mp}
\end{center}
\vspace{-2mm}
\end{figure}

Figure \ref{fig:blue_green_mp} shows scaling values for both model and model+data parallelism. We observe excellent scaling numbers in both settings. For example, the 8.3 billion parameters case with 8-way (8 GPU) model parallelism achieves 77\% of linear scaling. Model+data parallelism requires further communication of gradients and as a result the scaling numbers drop slightly. However, even for the largest configuration (8.3 billion parameters) running on 512 GPUs, we achieve 74\% scaling relative to linear scaling of the strong single GPU baseline configuration (1.2 billion parameters). Further scaling analysis is provided in Appendix \ref{app:scaling}

\subsection{Language Modeling Results Using GPT-2}
To demonstrate that large language models can further advance the state of the art, we consider training GPT-2 models of the sizes and configurations listed in Table \ref{tab:model_size}. The 355M model is equivalent in size and configuration of BERT-Large model \cite{devlin2018bert}. The 2.5B model is bigger than the previous largest GPT-2 model, and the 8.3B model is larger than any left-to-right transformer language model ever trained, to the best of our knowledge. To train and evaluate our language models we use the procedure described in section \ref{language_modeling}. Table \ref{tab:model_size} also lists the time it takes to advance one epoch which is equivalent to 68,507 iterations. For example, for the 8.3B model on 512 GPUs, each epoch takes around two days. Compared to the configurations used for our scaling studies in Table \ref{tab:params_scaling_studies}, the 2.5B model is the same, the 8.3B model has 24 attention heads instead of 32, and the 355M is much smaller than any seen previously while still using 64 GPUs to train, leading to the much lower time per epoch.

\begin{table}
\footnotesize
\begin{center}
\caption{Model configurations used for GPT-2.}
\label{tab:model_size}
\begin{tabular}{c@{\hskip3pt}|@{\hskip3pt}c@{\hskip3pt}|@{\hskip3pt}c@{\hskip3pt}|@{\hskip3pt}c@{\hskip3pt}|@{\hskip3pt}c@{\hskip3pt}|@{\hskip3pt}c@{\hskip3pt}|@{\hskip3pt}c} \hline \hline
 &  &  &  & Hidden &  & Time \\
Parameter & Layers & Hidden & Attn & Size & Total & per \\
Count &  & Size & Heads & per & GPUs & Epoch \\
 &  &  &  & Head &  & (days) \\ \hline
355M & 24 &  1024 &	16 & 64  &  64  & 0.86  \\ 
2.5B &  54 & 1920 & 20 & 96  &  128 & 2.27  \\ 
8.3B & 72 & 3072 &  24 &  128 & 512 & 2.10   \\ \hline 
\end{tabular}
\end{center}
\end{table}

\begin{figure}
\begin{center}
 \includegraphics[scale=0.19]{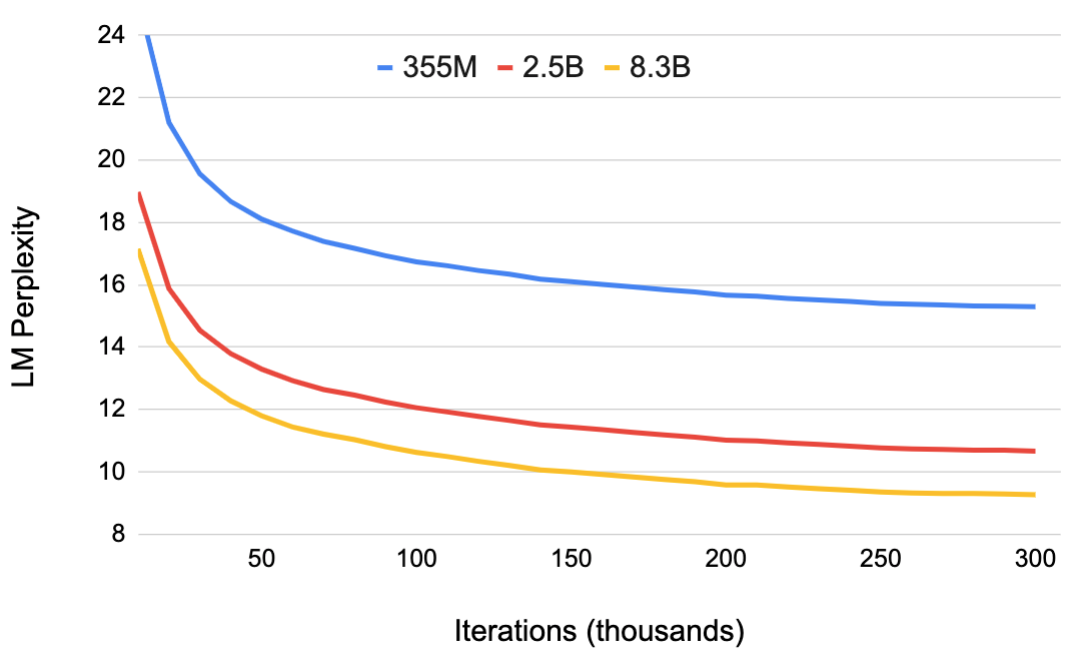}
 \caption{Validation set perplexity. All language models are trained for 300k iterations. Larger language models converge noticeably faster and converge to lower validation perplexities than their smaller counterparts.}
 \label{fig:ppl_curve}
\end{center}
\end{figure}

Figure \ref{fig:ppl_curve} shows validation perpelixity as a function of number of iterations. As the model size increases, the validation perpelixity decreases and reaches  a validation perplexity of 9.27 for the 8.3B model. We report the zero-shot evaluation of the trained models on the LAMBADA and WikiText103 datasets in Table \ref{tab:model_results}. For more details on evaluation methodology, see Appendix \ref{app:wikilambada}. We observe the trend that increasing model size also leads to lower perplexity on WikiText103 and higher cloze accuracy on LAMBADA. Our 8.3B model achieves state of the art perplexity on the WikiText103 test set at a properly adjusted perplexity of 10.81. At 66.51\% accuracy, the 8.3B model similarly surpasses prior cloze accuracy results on the LAMBADA task. We have included samples generated from the 8.3 billion parameters model in the Appendix \ref{app:samples}. Recently researchers from Microsoft in collaboration with NVIDIA trained a 17 billion parameter GPT-2 model called Turing-NLG \cite{TNLG} using Megatron and showed that the accuracies further improve as they scale the model, highlighting the value of larger models. 

\begin{table}
%\footnotesize
\begin{center}
\caption{Zero-shot results. SOTA are from \cite{WIKI103SOTA} for Wikitext103 and \cite{Radford2019GPT2} for LAMBADA.}
\label{tab:model_results}
\begin{tabular}{c|c|c} \hline \hline
Model & Wikitext103 & LAMBADA   \\
 & Perplexity $\downarrow$ & Accuracy $\uparrow$  \\ \hline
355M & 19.31 & 45.18\%  \\ 
2.5B & 12.76 & 61.73\%  \\ 
8.3B & \textbf{10.81} & \textbf{66.51\%}  \\ \hline 
Previous SOTA & 15.79 & 63.24\% \\
\end{tabular}
\end{center}
\end{table}
To ensure we do not train on any data found in our test sets, we calculate the percentage of test set 8-grams that also appear in our training set as done in previous work~\cite{Radford2019GPT2}. %To calculate the overlap we also use a Bloom filter but with a more conservative false positive rate of $10^{-3}$ to save computation cost.
The WikiText103 test set has at most $10.8\%$ overlap and the LAMBADA test set \citep{lambada} has at most $1.4\%$ overlap. We should note that the WikiText103 test set has already $9.09\%$ overlap with the WikiText103 training set \cite{Radford2019GPT2}. As these are consistent with previous work, 
we are confident that no documents from our test data are inadvertently included in our training data.

\begin{table}
\footnotesize
\begin{center}
\caption{Model configurations used for BERT.}
\label{tab:bert_model_size}
\begin{tabular}{c|c|c|c|c} \hline \hline
Parameter & Layers & Hidden & Attention & Total \\
Count     &        & Size   & Heads     & GPUs  \\ \hline
336M      & 24     & 1024   & 16        & 128   \\ 
1.3B      & 24     & 2048   & 32        & 256   \\ 
3.9B      & 48     & 2560   & 40        & 512   \\ \hline 
\end{tabular}
\end{center}
\end{table}

\begin{table*}
\footnotesize
\begin{center}
\caption{Development set results for MNLI, QQP, SQuAD 1.1 and SQuAD 2.0 and test set results for RACE. The trained tokens represents consumed tokens during model pretraining (proportional to batch size times number of iterations) normalized by consumed tokens during model pretraining for our 336M model.}
\label{tab:bert_results}
\begin{tabular}{c|c|c|c|c|c||c} \hline \hline
\multirow{3}{*}{Model}         & trained tokens & MNLI m/mm         & QQP       & SQuAD 1.1         & SQuAD 2.0        & RACE m/h         \\
              & ratio        & accuracy         & accuracy  & F1 / EM            & F1 / EM          & accuracy         \\
                  &        & (dev set)         & (dev set)  & (dev set)           & (dev set)          & (test set)         \\\hline
RoBERTa \cite{roberta}       & 2              & 90.2 / 90.2      & 92.2      & 94.6 / 88.9        & 89.4 / 86.5      & 83.2 (86.5 / 81.8)      \\
ALBERT \cite{ALBERT2019}        & 3              & 90.8             & 92.2      & 94.8 / 89.3        & 90.2 / 87.4      & 86.5 (89.0 / 85.5)      \\
XLNet \cite{xlnet}       & 2              & 90.8 / 90.8      & 92.3      & 95.1 / 89.7   & 90.6 / 87.9      & 85.4 (88.6 / 84.0)      \\
Megatron-336M & 1              & 89.7 / 90.0      & 92.3      & 94.2 / 88.0        & 88.1 / 84.8      & 83.0 (86.9 / 81.5)      \\
Megatron-1.3B & 1              & 90.9 / 91.0      & 92.6      & 94.9 / 89.1        & 90.2 / 87.1      & 87.3 (90.4 / 86.1)      \\
Megatron-3.9B & 1           & \bf{91.4 / 91.4} & \bf{92.7} & \bf{95.5 / 90.0} & \bf{91.2 / 88.5} & \bf{89.5 (91.8 / 88.6)} \\\hline
 \multicolumn{3}{l}{ALBERT ensemble \cite{ALBERT2019} } & & 95.5 / 90.1 &91.4 / 88.9 & 89.4 (91.2 / 88.6)\\
  \multicolumn{3}{l}{Megatron-3.9B ensemble} &  &  \bf{95.8 / 90.5} & \bf{91.7 / 89.0} & \bf{90.9 (93.1 / 90.0)}\\\hline 
\end{tabular}
\end{center}
\end{table*}

\subsection{Bi-directional Transformer Results Using BERT}

In this section, we apply our methodology to BERT-style transformer models and study the effect of model scaling on several downstream tasks. Prior work \cite{ALBERT2019} found that increasing model size beyond BERT-large with 336M parameters results in unexpected model degradation. To address this degradation, the authors of that work \cite{ALBERT2019} introduced parameter sharing and showed that that their models scale much better compared to the original BERT model.

\begin{figure}
\vspace{-2mm}
\begin{center}
  \includegraphics[scale=0.13]{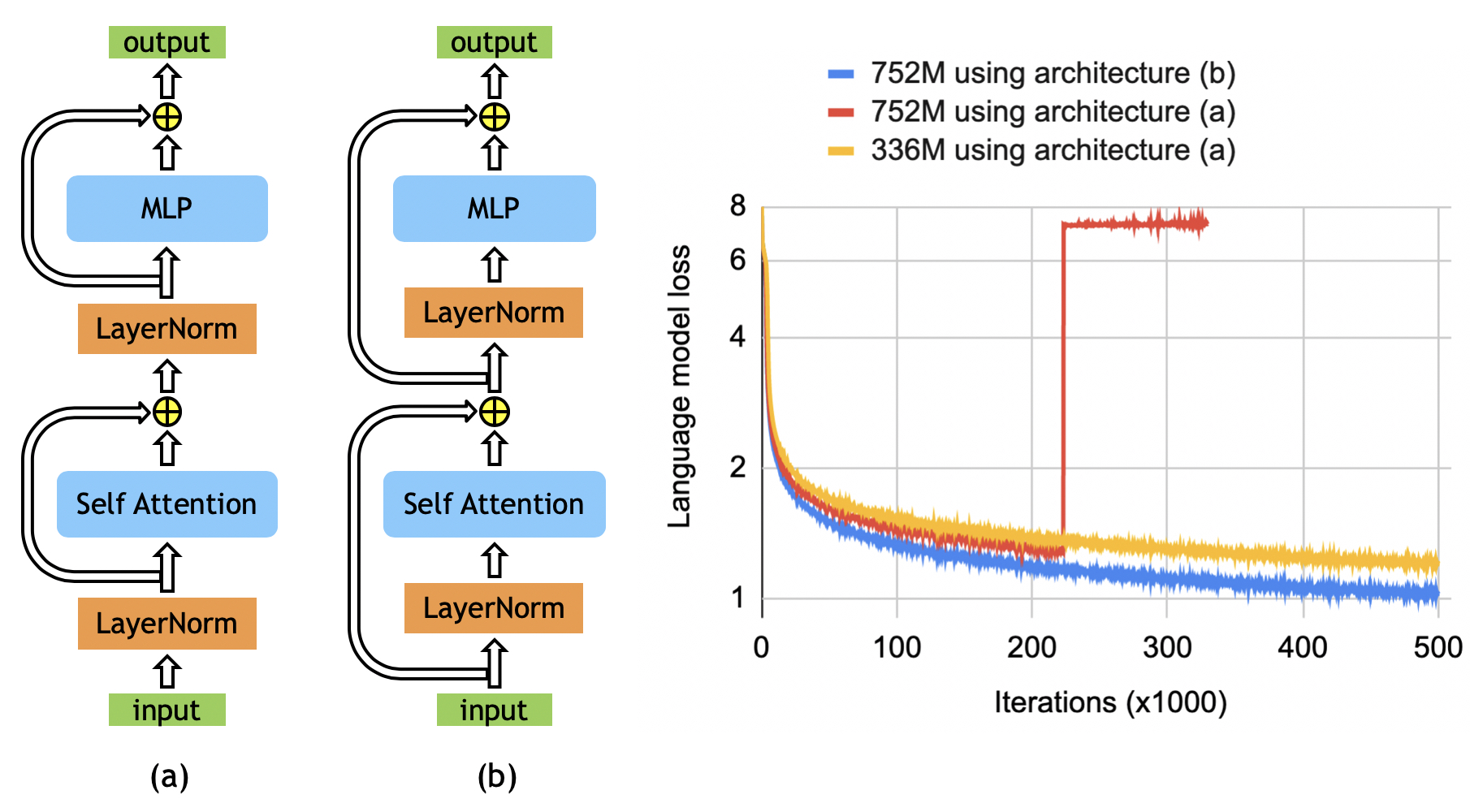}
  \vspace{-2mm}
  \caption{Training loss for BERT model using the original architecture (a) and the rearranged architecture (b). Left figure shows the training loss for 336M and 752M BERT model. While the original architecture performs well on the 336M model, the modifications in (b) enable stable training with lower training loss.}
  \label{fig:bert-residual}
\end{center}
\vspace{-6mm}
\end{figure}

We further investigated this behaviour and empirically demonstrated that rearranging the order of the layer normalization and the residual connections as shown in Figure \ref{fig:bert-residual} is critical to enable the scaling of the BERT-style models beyond BERT-Large. The architecture (b) in Figure \ref{fig:bert-residual} eliminates instabilities observed using the original BERT architecture in (a) and also has a lower training loss. To the best of our knowledge, we are the first to report such a change enables training larger BERT models. 

Using the architecture change in Figure~\ref{fig:bert-residual}(b), we consider three different cases as detailed in Table~\ref{tab:bert_model_size}. The 336M model has the same size as BERT-large. The 1.3B is the same as the BERT-xlarge configuration that was previously shown to get worse results than the 336M BERT-large model \cite{ALBERT2019}. We further scale the BERT model using both larger hidden size as well as more layers to arrive at the 3.9B parameter case. In all cases, the hidden size per attention head is kept constant at 64. 336M and 1.3B models are trained for 2 million iterations while the 3.9B model is trained for 1.5 million iterations and is still training.

On a 3\% held-out set, 336M, 1.3B, and 3.9B models achieve validation set perplexity of 1.58, 1.30, and 1.16, respectively, a monotonic decrease with the model size. We finetune the trained models on several downstream tasks including MNLI and QQP from the GLUE benchmark \cite{GLUE2019}, SQuAD 1.1 and SQuAD 2.0 from the Stanford Question answering dataset \cite{SQUAD1,SQUAD2}, and the reading comprehension RACE dataset \cite{RACE}. For finetuning, we follow the same procedure as \cite{roberta}. We first perform hyperparameter tuning on batch size and learning rate. Once we obtain the best values, we report the median development set results over 5 different random seeds for initialization. The hyperparameters used for each model and task are provided in the Appendix \ref{app:berthparams}. Table \ref{tab:bert_results} shows the development set results for MNLI, QQP, SQuAD 1.1, and SQuAD 2.0 and test set results for RACE. For the test set results of RACE, we first use the development set to find the checkpoint that gives us the median score on the 5 random seeds and we report the results from that checkpoint on the test set. We also report 5-way ensemble results for the development set of SQuAD and test set of RACE. From Table \ref{tab:bert_results} we observe that (a) as the model size increases, the downstream task performance improves in all cases, (b) our 3.9B model establishes state of the art results on the development set compared to other BERT based models, and (c) our 3.9B model achieves both single model as well as ensembled  SOTA results on RACE test set.

\section{Conclusion and Future Work}

In this work, we successfully surpassed the limitations posed by traditional single-GPU-per-model training by implementing model parallelism with only a few modifications to the existing PyTorch transformer implementations. We efficiently trained transformer based models up to 8.3 billion parameter on 512 NVIDIA V100 GPUs with 8-way model parallelism and achieved up to 15.1 PetaFLOPs sustained over the entire application. We also showed that for BERT models, careful attention to the placement of layer normalization in BERT-like models is critical to achieving increased accuracies as the model size increases. We study the effect of model size on down-stream task accuracy and achieve far superior results on downstream tasks and establish new SOTA for WikiText103, LAMBADA, and RACE datasets. Finally, we open sourced our code to enable future work leveraging model parallel transformers.

There are several directions for future work. Continuing to increase the scale of pretraining is a promising line of investigation that will further test existing deep learning hardware and software. To realize this, improvements in the efficiency and memory footprint of optimizers will be needed. In addition, training a model with more than 16 billion parameters will demand more memory than is available within 16 GPUs of a DGX-2H box. For such models, a hybrid intra-layer and inter-layer model parallelism along with inter-node model parallelism would be more suitable.
%Increasing the scale of pretraining and transfer is not the only way to demonstrate the effectiveness of large scale language modeling. To this end 
Three other directions of investigation include (a) pretraining different model families (XLNet, T5), (b) evaluating performance of large models across more difficult and diverse downstream tasks (e.g. Generative Question Answering, Summarization, and Conversation), and (c) using knowledge distillation to train small student models from these large pretrained teacher models.

\Urlmuskip=0mu plus 1mu\relax
\bibliography{main}
\bibliographystyle{icml2020}

\appendix
%\documentclass{article}
%\usepackage{graphicx}
%\usepackage{amsmath}
%\usepackage{icml2020}
%\twocolumn

%\begin{document}

\section{BERT Finetuning Hyperparameters}
\label{app:berthparams}

Table \ref{tab:bert_tasks_hparams} presents the hyperparameters used for each model and task during finetuning.

\begin{table}[hbt!]
\footnotesize
\begin{center}
\caption{Hyperparameters for finetuning BERT model on downstream tasks.}
\label{tab:bert_tasks_hparams}
\begin{tabular}{c|c|c|c|c} \hline \hline
Task      & Model & Batch & Learning & Training \\
          &       & size  & rate     & epochs   \\ \hline
          & 336M  &       &          &          \\ 
MNLI      & 1.3B  & 128   & 1e-5     & 10       \\ 
          & 3.8B  &       &          &          \\ \hline 
          & 336M  &   128    & 5e-5     &          \\ 
QQP       & 1.3B  & 128   & 3e-5     & 12      \\ 
          & 3.8B  &  256     & 4e-5     &          \\ \hline 
          & 336M  & 64    & 3e-5     &          \\ 
SQUAD 1.1 & 1.3B  & 48    & 3e-5     & 2        \\ 
          & 3.8B  & 48    & 1e-5     &          \\ \hline 
          & 336M  & 48    & 3e-5     &          \\ 
SQUAD 2.0 & 1.3B  & 64    & 3e-5     & 2        \\ 
          & 3.8B  & 48    & 1e-5     &          \\ \hline 
          & 336M  & 32    & 2e-5     &          \\ 
RACE      & 1.3B  & 16   & 1e-5     & 3        \\ 
          & 3.8B  & 32    & 2e-5     &          \\ \hline 
\end{tabular}
\end{center}
\end{table}

\section{Model Parallel Supplementary Material }
\label{sec:modelpar:supp}

In this section, we present further details about the hybrid model and data parallelism and handling random number generation.

\subsection{Hybrid Model and Data Parallelism}
\begin{figure}[h]
\begin{center}
 \includegraphics[scale=0.32]{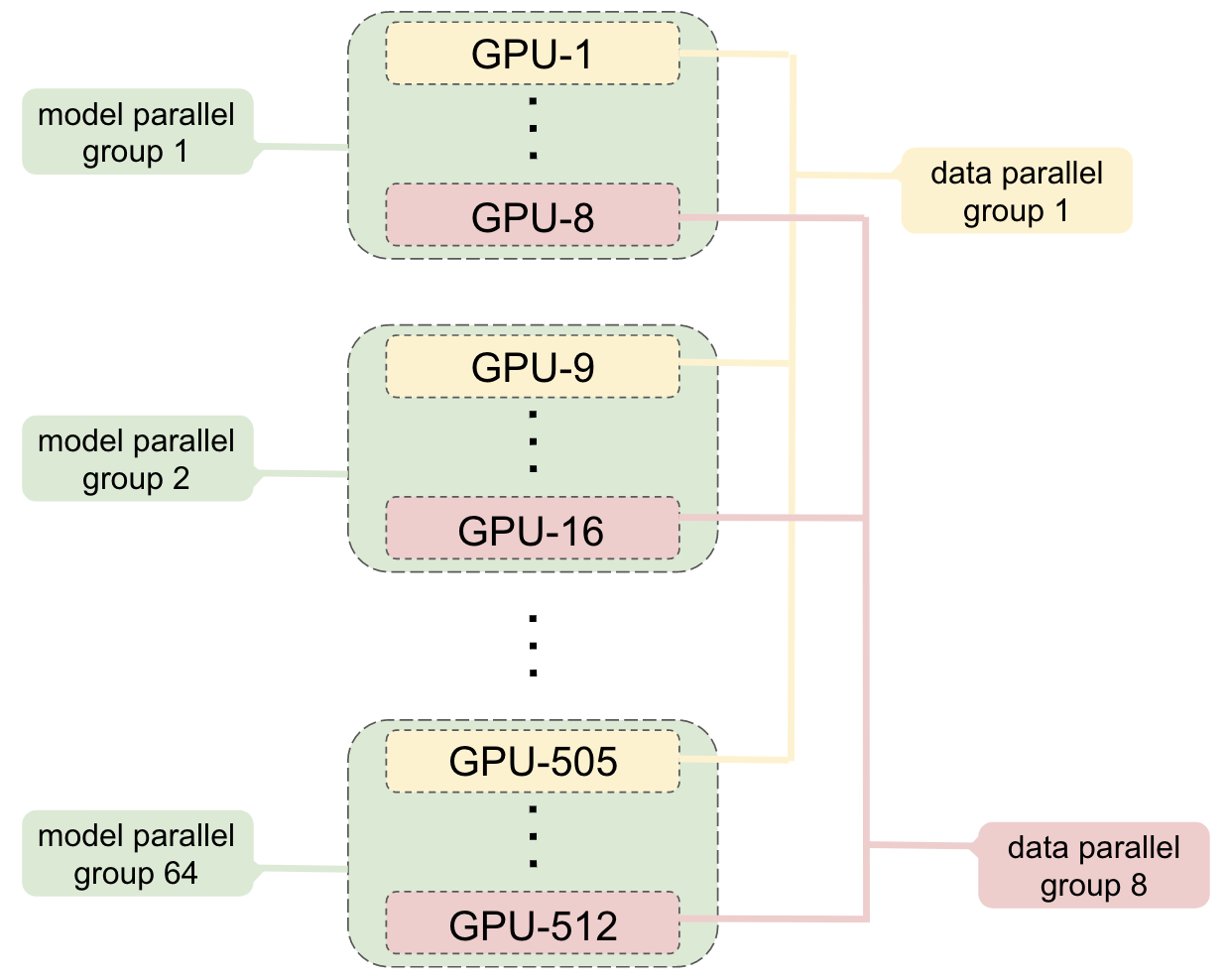}
 \caption{Grouping of GPUs for hybrid model and data parallelism with 8-way model parallel and 64-way data parallel.}
 \label{fig:hybrid-mpdp}
\end{center}
\vspace{-6mm}
\end{figure}

Model parallelism is orthogonal to data parallelism, and so we can use both simultaneously to train large models in a reasonable amount of time. Figure \ref{fig:hybrid-mpdp} shows a grouping of GPUs for hybrid model and data parallelism. Two or more GPUs within the same server form model parallel groups (for example GPUs 1 to 8 in Figure \ref{fig:hybrid-mpdp}), and contain one instance of the model distributed across these GPUs.  The remaining GPUs, which could be within the same server but more typically are located in other servers, run additional model parallel groups. GPUs with the same position in each of the model parallel groups (for example GPUs 1, 9, ..., 505 in Figure \ref{fig:hybrid-mpdp}) form data parallel groups so that all GPUs within a data parallel group hold the same model parameters. During back propagation we run multiple gradient all-reduce operations in parallel to reduce weight gradients within each distinct data parallel group. The total number of required GPUs is the product of the number of model and data parallel groups. For example, for the 8.3 billion parameter model we use 8 GPUs per model parallel group and 64-way data parallelism, for a total of 512 GPUs.  All communication is implemented in PyTorch by Python calls to NCCL.  GPUs within each model parallel group perform all-reduces amongst all GPUs within the group.  For data parallelism, each of the all-reduce operations takes place with one of the GPUs from each model parallel group.

\subsection{Model Parallel Random Number Generation}
Techniques that utilize random number generation, such as dropout, are a staple of modern deep learning training. Transformers have dropout layers outside the model parallel regions before residual connections and within model parallel regions in the self attention block. Because some dropout layers are in a model parallel region, while others are not, we need to treat random number generation carefully to ensure dropout works correctly. To synchronize residual connection dropout across model parallel workers we seed the random number generators at the beginning of training with the same seed. This results in identical dropout patterns across all model parallel workers. However, dropout within a model parallel region should result in different random patterns for each worker to achieve randomness across the entire operation. To achieve this we maintain a separate random number generator for dropout within model parallel regions. This random number generator is uniquely seeded for each model parallel worker.

\section{Text Samples}
\label{app:samples}
Below are some text samples generated by Megatron-LM using a context prompt. Some of the texts are cut short.
\fbox{
\parbox{0.48\textwidth}{
\rule{0.48\textwidth}{0.4pt}
{\bf Context}: There is a major force uniting America’s fiercely partisan politicians: big technology companies. Democrats and Republicans at the federal and state levels are coming together to scrutinize the power of the Silicon Valley giants and, potentially, to rein them in. Letitia James, the Democratic attorney general of New York, announced on Friday that attorneys general in eight states — four Democrats and four Republicans — and the District of Columbia had begun an antitrust investigation of Facebook.

\rule{0.48\textwidth}{0.4pt}
{\bf Megatron-LM}: The probe, which Ms. James said was likely to result in a lawsuit, is the latest in a string of high-profile regulatory actions, in state and federal courts, that challenge the market dominance of tech companies. The Justice Department is also investigating Facebook's data practices and whether the company violated a 2011 consent decree with the Federal Trade Commission that required the company to obtain explicit consent from users before sharing their personal information with third parties.
\\[1\baselineskip]
What might prompt a bipartisan coalition to sue Facebook? It isn't clear that there's any one thing. But lawmakers and regulators have been increasingly alarmed by the company's recent controversies, which include a data breach affecting more than 50 million users, the spread of fake news on the platform, and the spreading of propaganda by Russian-backed accounts during the 2016 election.
\\[1\baselineskip]
It's also not clear how serious the Democratic attorneys general are about filing a lawsuit. While the attorneys general announced their investigation in a press release, they said they had not yet reached a decision on whether to sue. In the past, some of the attorneys general have focused their efforts on trying to extract settlements from the tech companies without filing lawsuits.
\\[1\baselineskip]
"As attorneys general who have investigated and prosecuted health-care fraud, price-fixing, and other violations of our states' antitrust laws, we are committed to holding companies accountable when they fail to protect consumers and competition in the market," Ms. James said in a statement.
\\[1\baselineskip]
The investigation, which was first reported by The New York Times, was announced as the companies that dominate the tech industry are facing scrutiny from antitrust regulators around the world. Google, Twitter, Facebook and Amazon have faced a series of antitrust probes and class-action lawsuits in the United States and Europe ... 
%Last month, European regulators fined Google \$2.7 billion for unfairly promoting its own shopping service. And the European Union's competition commissioner, Margrethe Vestager, has threatened to hit Google with additional penalties.

}}

\fbox{%
\parbox{0.48\textwidth}{%
\rule{0.48\textwidth}{0.4pt}
{\bf Context}: A flower, sometimes known as a bloom or blossom, is the reproductive structure found in flowering plants (plants of the division Magnoliophyta, also called angiosperms). The biological function of a flower is to effect reproduction, usually by providing a mechanism for the union of sperm with eggs.

\rule{0.48\textwidth}{0.4pt}
{\bf Megatron-LM}: The structural components of the flower are the petals and sepals, which are the reproductive organs, the hypanthium, which forms the base of the flower, and the calyx, which is the stigma. A flower is a composite structure, made up of sepals and petals, the sepals being composed of three fused tepals and the petals of two fused petals, though there are a few exceptions to this rule. The three sepals are fused to form a tube called the calyx, which is the structure that attracts the pollinating insects. The petals are composed of two fused sepals (the outer ones) which open and fold inward (clasp) at the tips, forming the petals. The four sepals and petals are united to form a corolla, the outermost structure of a flower that carries pollen.
\\[1\baselineskip]
The structure of the flower has important functional significance, including protection of the ovary from insects and the transfer of pollen between the male and female parts of a flower. The structure of the flower is also used by the plants to attract pollinators, for example by the shape of the petals and the colours of the stamens.
\\[1\baselineskip]
The structure of a flower varies widely between plants. The simplest flowers are unbranched and made up of only two sepals and a single petal. Other examples are composed of several petals joined together, with the innermost sepals completely fused to form a tube called the calyx. The petals are usually coloured to attract insects, which can only see certain wavelengths of light. The sepals are often coloured too, but usually have a more limited range of colours.
\\[1\baselineskip]
Some flowers have additional structures, such as stamens, anthers and ovaries. The presence of stamens and anthers in some flowers is used for cross-pollination. The ovary is the organ where the female gametes are produced, and is the first structure that is developed in an embryo of a flower. The ovary is the female flower's reproductive system, and consists of the ovary wall, the ovule, and the oviduct.
\\[1\baselineskip]
Flowers vary in their size and shape, from tiny to huge, and are often brightly coloured. Flowers may be seen as part of a larger structure, such as a flower head, an inflorescence, a bract, a spathe, or a carpels.
\\[1\baselineskip]
The term "flower" is usually applied to the floral structures that are visible externally, but many ...
%flowers produce their flowers internally, and these structures are collectively known as the "floral anatomy" or "floral anatomy of a plant". The floral anatomy includes structures that are only visible internally, such as the ovary, the style, and the stigma.
%\newline \newline
%The term flower is also applied to the fruit of a flowering plant, such as the berry of a strawberry. The term may also be used to refer to the various parts
%of the flower that are not visible externally.
%The term "flower" is also used in other contexts, including the common names of certain plants and the eponymous genus of an entire family of flowering plants, Asteraceae.

}}

\fbox{%
\parbox{0.48\textwidth}{%
\rule{0.48\textwidth}{0.4pt}
{\bf Context}: OpenAI did not release the full GPT-2 model due to concerns of malicious use, but they did release a smaller version equivalent in size to the original GPT (117 M parameters), trained on the new, larger dataset. Although not as powerful as the large model, the smaller version still has some language generation chops.

\rule{0.48\textwidth}{0.4pt}
{\bf Megatron-LM}: In December 2016, DeepMind released an even larger model, GPT-3, with more than 1.5 billion parameters. DeepMind has no plans to release the model, so the question remains whether other AI researchers will be able to copy and build on it.
\\[1\baselineskip]
Other language models have been built using other techniques. In 2016, a team of researchers at Columbia University, the University of Toronto, and Google DeepMind showed that neural network-based models, including deep neural networks, can learn the grammatical rules of a language. A group at Google DeepMind demonstrated that these rules could be represented in a network as a kind of decision tree, where the nodes in the tree are the words in the sentence, and the rules for the words are represented by the decision nodes at the end of the tree. The network would go through the network in order to choose the next word in the sentence based on whether it follows the rules or not. In 2017, the team at DeepMind released a paper detailing the training and performance of their model.
\\[1\baselineskip]
The problems with the model are many. One of the biggest issues is that grammatical rules are ambiguous. For example, the phrase "I came down" could mean I came down from the stairs, or I came down from the sky. In English, there are numerous other examples of sentences that are ambiguous. The model does not have any way to tell the difference between the ambiguous sentences, and so it simply makes a prediction about the entire sentence based on what rules it does know. This makes the model vulnerable to the phenomenon of overfitting. In order to avoid this, models must be trained using a large set of data, so that the model can find useful patterns in the data, and use those patterns to help make predictions. This approach is also prone to errors, as it relies on the training data to help make correct predictions, and the data might be incorrectly labeled, resulting in a model that produces a poor set of predictions. In the case of GPT-2, the model only has about 100 training sentences for each type of phrase (subject-verb-object).
\\[1\baselineskip]
Additionally, languages have different rules for different types of phrasing. In English, the subject-verb-object construction is fairly universal, but other languages have different rules, which makes training a GPT more complicated.
\\[1\baselineskip]
Finally, there is an inherent trade-off between the number of rules and the performance of the model ... 
%As a rule, the more rules the model has, the better it performs. However, as the number of rules increases, the likelihood of the model to overfit the training data increases, and the ability to perform generalizations decreases. For instance, if the model has 1,000 rules ... 
%for an English phrase, it will be able to predict other English phrases with high accuracy. However, it will struggle with generalizing to other languages. If the model has 10,000 rules, it will be able to perform generalizations, but it will be inaccurate with respect to English.
%There have been other grammar learning algorithms that have not yet been released to the public. One such system was used to produce a language translation system that was able to translate into and out of English. However, the system did not learn the rules for the different languages, but instead relied on information about the structure of the language to translate between the two languages. Another system has also been developed by Google DeepMind, which used neural networks to learn grammar. The system was able to learn grammatical rules for two languages, English and French.

}}

\fbox{%
\parbox{0.48\textwidth}{%
\rule{0.48\textwidth}{0.4pt}
{\bf Context}: The 36th International Conference on Machine Learning (ICML 2019) will be held in Long Beach, CA, USA from June 10th to June 15th, 2019. The conference will consist of one day of tutorials (June 10), followed by three days of main conference sessions (June 11-13), followed by two days of workshops (June 14-15).

\rule{0.48\textwidth}{0.4pt}
{\bf Megatron-LM}: With a broad scope, the conference addresses the challenges and opportunities in machine learning for practitioners and researchers. The conference program will feature sessions that cover different machine learning topics, from reinforcement learning to computational biology and from AI to deep learning. It is the world's premier forum for exploring and presenting advances in machine learning.
\\[1\baselineskip]
ICML 2019 will offer an impressive range of sessions, which will be of interest to practitioners, researchers, and students. Some of the topics we will cover include:
\\[1\baselineskip]
Reinforcement learning
\\[1\baselineskip]
Soft Computing
\\[1\baselineskip]
Computational biology
\\[1\baselineskip]
Convolutional networks
\\[1\baselineskip]
Deep learning
\\[1\baselineskip]
Statistical learning theory
\\[1\baselineskip]
Statistical learning machine
\\[1\baselineskip]
Natural language processing
\\[1\baselineskip]
Biology
\\[1\baselineskip]
Awards
\\[1\baselineskip]
Since its first edition, the conference has grown and evolved. Today, it is the premier forum for showcasing state-of-the-art research in machine learning. The conference is widely recognized as the preeminent venue for showcasing the latest research in machine learning, and a place where practitioners, researchers, and students can exchange ideas and insights.
\\[1\baselineskip]
The conference is a peer-reviewed venue that draws its prestige and influence from the excellence of the presentations and the quality of the invited speakers. With over 50\% of all talks selected based on their content by an expert committee, it is clear that the conference's high standards of research and quality keep a very high bar.
%\newline \newline
%For more information about the conference, please visit: ICML.
}}

\fbox{%
\parbox{0.48\textwidth}{%
\rule{0.48\textwidth}{0.4pt}
{\bf Context}: Rescue workers on Tuesday found a flight recorder from the Russian plane that crashed into the Black Sea over the weekend, the defense ministry said. All 92 people aboard the Russian military's Tu-154 plane are believed to have died Sunday morning when it crashed two minutes after taking off from the southern Russian city of Sochi.

\rule{0.48\textwidth}{0.4pt}
{\bf Megatron-LM}: A child holds up a portrait of a dead relative at a makeshift morgue in a building belonging to a local government in the town of Petropavlovsk-Kamchatsky. A Russian Defense Ministry Tu-154 plane with 92 people on board crashed into the Black Sea on Sunday morning shortly after taking off from the southern Russian city of Sochi. (Yuri Kochetkov/EPA)
\\[1\baselineskip]
According to the ministry, the plane was found at a depth of more than 1,000 feet and the recorder was damaged. "The defense ministry has decided to hand over the recorder to the Interstate Aviation Committee, which is the main investigative body in the air crash," the statement said.
\\[1\baselineskip]
The search for the plane's black boxes is being hampered by strong winds and rough seas. Investigators have launched a criminal probe into the crash, which officials said was likely caused by a technical fault.
\\[1\baselineskip]
Russian President Vladimir Putin declared Wednesday a day of mourning for the victims.

}}

\section{Further Scaling Analysis}
\label{app:scaling}

In this section we study the effect of number of attention heads on the scaling results. We also present strong scaling results for our 1.2 billion parameter model.

\subsection{Attention Heads and Scaling} 
This section studies the effect of attention heads on model parallel scaling. To this end, we consider the 8.3 billion parameter configuration with 8-way model parallelism and vary the number of heads from 16 to 32. The results are presented in Table \ref{tab:attn_head_scaling}. As the number of attention heads increases, some of the GEMMS inside the self-attention layer become smaller and also the number of elements in the self attention softmax increases. This results in a slight decrease in scaling efficiency. Future research should be wary of this hyperparameter to design large transformer models that balance model speed and model accuracy.

\begin{table}[hbt!]
\footnotesize
\begin{center}
\caption{Effect of number of attention heads on scaling on 8.3 billion of parameters with 8-way model parallelism.}
\vskip 0.15in
\label{tab:attn_head_scaling}
\begin{tabular}{c|c|c}
\hline \hline
Attention heads & Hidden size per head & Scaling Efficiency \\ \hline
16 & 192 & 82\% \\
24 & 128 & 80\% \\
32 & 96 & 77\% \\ \hline \hline
\end{tabular}
\end{center}
\vskip -0.1in
\end{table}

\subsection{Strong Scaling}

Our model parallelism is primarily designed to enable training models larger than what can fit in the memory of a single GPU, but it can also accelerate the training of smaller models without increasing the batch size. To measure this acceleration we train a model with a fixed 1.2 billion parameters. We use a fixed batch size of 8 samples per iteration and increase the number of GPUs using model parallelism. The results are listed in Table~\ref{tab:strong_scaling}. Using two GPUs makes training $64\%$ faster. Above that we see diminishing returns as the per-GPU computation decreases and the memory bandwidth and communication overheads begin to dominate.

\begin{table}
\centering
\caption{Speedup obtained for the 1.2 billion parameters model using model parallelism while keeping the batch size constant.}
\vskip 0.15in
\begin{tabular}{c|c|c|c|c}
\hline \hline
\# of GPUs & 1 & 2 & 4 & 8 \\
\hline
Speedup & 1.0 & 1.64 & 2.34 & 2.98 \\
\hline \hline
\end{tabular}
\label{tab:strong_scaling}
\vskip -0.1in
\end{table}

\section{Evaluating Language Models Using WikiText103 and LAMBADA}
\label{app:wikilambada}
In this section we detail our evaluation methodology for the WikiText103 dataset \citep{wikitext} and cloze-style prediction accuracy on the LAMBADA dataset\citep{lambada}.

\subsection{Wikitext103 Perplexity}
WikiText103 perplexity is an evaluation criterion that has been well studied over the past few years since the creation of the benchmark dataset. Perplexity is the exponentiation of the average cross entropy of a corpus \citep{perplexity}. This makes it a natural evaluation metric for language models which represent a probability distribution over entire sentences or texts. 

\begin{equation}
 PPL= \exp({-\frac{1}{T_o}\sum_{t}^{T} \text{log} P(t|0:t-1))}
 \label{ppl_eqn}
\end{equation}

To calculate perplexity in (\ref{ppl_eqn}) we tokenize the WikiText103 test corpus according to our subword vocabulary and sum the cross entropy loss from each token $[0, T]$. We then normalize the cross entropy loss by the number of tokens in the original tokenization scheme $T_o$. The WikiText103 test corpus already comes pre-tokenized with word level tokens that prior works have used to compute perplexity. To evaluate our models' perplexities on a level playing field with prior works we must normalize by the original number of tokens, $T_o$, rather than the number of tokens, $T$, actually in the tokenized data fed as input to our model.  This pre-tokenization also introduces artifacts in the text that are not present in our training data. To alleviate this distributional mismatch, we first preprocess the WikiText103 test dataset with invertible detokenizers to remove various artifacts related to punctuation and whitespace. The value of $T_o$ is calculated before this preprocessing. For WikiText103's test set $T_o=245566$ and $T=270329$.

We must also make one further transformer-specific modification to the perplexity calculation. Unlike RNN-based language models, transformers operate on a fixed window input size. Therefore they cannot fully calculate $P(t|0:t-1)$ and can only calculate $P(t|t-w:t-1)$ where $w$ is the size of our context: 1024 tokens. However, calculating this value for every token in our dataset is prohibitively expensive since we must compute approximately $T$ evaluations of a $w$ sized context. To evaluate our models efficiently we take a middle ground approach termed \textit{overlapping evaluation} where we advance the sliding window by some overlap $o$ each time and only compute the cross entropy losses corresponding to the last $o$ tokens of the window. In our experiments we utilize an overlap $o$ of 32, and compute losses over all sliding windows in such a fashion.

\subsection{LAMBADA Cloze Accuracy}
The capability to handle long term contexts is crucial for state of the art language models and is a necessary prerequisite for problems like long-form generation and
document-based question answering. Cloze-style datasets like LAMBADA are designed to measure a model's ability to operate in and reason about these types of long term contexts. Cloze-style reading comprehension uses a context of word tokens $x=x_{1:t}$ with one token $x_j$ masked; the models objective is to correctly predict the value of the missing $j^{\text{th}}$ token. To accurately predict the missing token, the model requires an in-depth understanding of the surrounding context and how language should be used in such a context. LAMBADA uses cloze-style reading comprehension to test generative left-to-right language models by constructing examples of 4-5 sentences where the last word in the context $x_t$ is masked. Our models utilize subword units, so for LAMBADA evaluation we utilize the raw, unprocessed LAMBADA dataset and require that our model predict the multiple subword tokens that make up the word token. We use teacher forcing, and consider an answer correct only when all output predictions are correct. This formulation is equivalent to the original task of word token prediction.

%\bibliography{megatron}
%\bibliographystyle{icml2020}

%\end{document}

\end{document}